\documentclass[letterpaper, 10 pt, conference]{ieeeconf}

\IEEEoverridecommandlockouts

\usepackage{amsmath}
\usepackage{mathtools}
\usepackage{booktabs}
\usepackage{float}
\usepackage{xspace}
\usepackage{cite}
\usepackage{graphicx}
\graphicspath{{./figures/}}
\usepackage{caption}
\usepackage{subcaption}
\usepackage[inline]{asymptote}
\usepackage{comment}
\usepackage[dvipsnames]{xcolor}
\usepackage{hyperref}
\usepackage{babel}

\usepackage{multirow}

\usepackage{makecell}

\usepackage{caption,tabularx,booktabs}

\usepackage[linesnumbered,ruled,vlined]{algorithm2e}
\usepackage{algpseudocode}

\definecolor{deeppink}{rgb}{1.0, 0.08, 0.58}
\hypersetup{
  colorlinks = true, %
  urlcolor   = deeppink, %
  linkcolor  = blue, %
  citecolor  = red %
}

\usepackage{graphicx}

\PassOptionsToPackage{hyphens}{url}

\usepackage{array}
\newcommand{\PreserveBackslash}[1]{\let\temp=\\#1\let\\=\temp}
\newcolumntype{C}[1]{>{\PreserveBackslash\centering}p{#1}}
\newcolumntype{L}[1]{>{\PreserveBackslash\raggedright}p{#1}}

\newcommand{\website}{https://cmu-mfi.github.io/plc-safety/}

\title{\LARGE \bf
Robot Safety Monitoring using Programmable Light Curtains
}

\begin{document}

\global\csname @topnum\endcsname 0
\global\csname @botnum\endcsname 0

\bstctlcite{Force_Etal}
\definecolor{cadmiumred}{rgb}{0.89, 0.0, 0.13}

\makeatletter
\let\@oldmaketitle\@maketitle
\renewcommand{\@maketitle}{\@oldmaketitle
    \centerline{
    \hspace*{-0.2em}
    \begin{minipage}[t]{0.30\linewidth}
        \centering
        \includegraphics[height=130px]{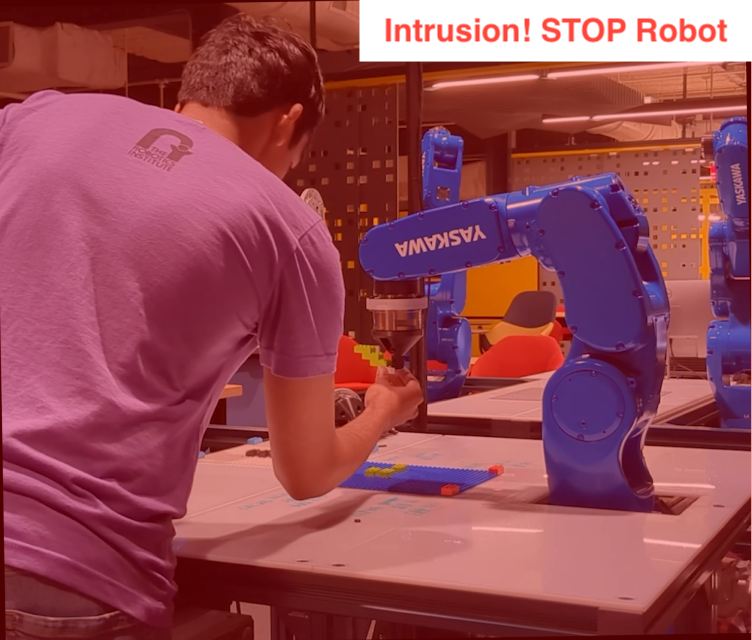}
        \captionof*{figure}{(a) Robot workspace}
    \end{minipage}%
    \hspace*{-0.9em}
    \begin{minipage}[t]{0.30\linewidth}
        \centering
        \includegraphics[height=130px,trim={0.0cm 1cm 0.5cm 0.5cm},clip,]{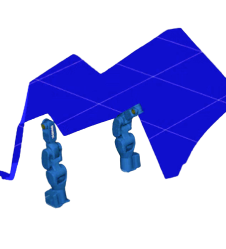}
        \captionof*{figure}{(b) Light curtain envelope}
    \end{minipage}%
    \hspace*{-1em}
    \begin{minipage}[t]{0.25\linewidth}
        \centering
        \includegraphics[height=130px]{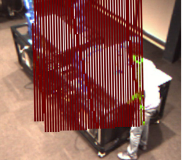}
        \captionof*{figure}{(c) Light curtain overlay}
    \end{minipage}%
    \hspace*{1.1em}
    \begin{minipage}[t]{0.15\linewidth}
        \centering
        \includegraphics[height=130px]{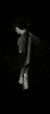}
        \captionof*{figure}{(b) Intensity image}
    \end{minipage}%
}
\captionof{figure}{
    \textbf{Overview}. {\bf(a)} A human worker walks into a robot workspace to adjust the work-piece on its end-effector. {\bf(b)} A programmble light curtain (PLC) forms a tight convex hull enveloping the two robots. This acts as a flexible ``safety shield'' for collision monitoring that adapts to the robot's motion in real-time. {\bf(c)} The intrusion detected  by 3D points on the light curtain intersecting the obstacle are shown in green. This triggers the robot to stop. {\bf(d)} The full intensity image captured by the PLC shows the outline of the detected worker.
}
    \label{fig:teaser}
} 
\makeatother

\author{Karnik Ram$^{1\to2}$, Shobhit Aggarwal$^1$, Robert Tamburo$^1$, Siddharth Ancha$^3$, Srinivasa Narasimhan$^1$
\\\\
{
\vspace*{0cm}
\normalsize
Supplementary website: \href{\website}{\texttt{\website}}}$^*$
\vspace{0.5em}
\thanks{$1$\;Robotics Institute at Carnegie Mellon University, Pittsburgh, USA. $2$\;TU Munich, Germany. $3$\;MIT, Cambridge, USA. Correspondence email: \texttt{karnikram@gmail.com}, \texttt{\{shobhita, srinivas\}@andrew.cmu.edu}. $^*$While this paper is fully self-contained, our \href{\website}{project website} contains an overview video and qualitative visualizations for easy access.}\\
}

\maketitle
\thispagestyle{empty}
\pagestyle{empty}

\begin{abstract}
    As factories continue to evolve into collaborative spaces with multiple robots working together with human supervisors in the loop, ensuring safety for all actors involved becomes critical. Currently, laser-based light curtain sensors are widely used in factories for safety monitoring. While these conventional safety sensors meet high accuracy standards, they are difficult to reconfigure and can only monitor a fixed user-defined region of space. Furthermore, they are typically expensive. Instead, we leverage a controllable depth sensor, \textit{programmable light curtains} (PLC), to develop an inexpensive and flexible real-time safety monitoring system for collaborative robot workspaces. Our system projects virtual \textit{dynamic safety envelopes} that tightly envelop the moving robot at all times and detect any objects that intrude the envelope. Furthermore, we develop an instrumentation algorithm that optimally places (multiple) PLCs in a workspace to maximize the visibility coverage of robots. Our work enables fence-less human-robot collaboration, while scaling to monitor multiple robots with few sensors. We analyze our system in a real manufacturing testbed with four robot arms and demonstrate its capabilities as a fast, accurate, and inexpensive safety monitoring solution.

\end{abstract}

\begin{figure*}[t!]
    \centerline{
    \subfloat[\small PLC prototype]{
        \includegraphics[width=0.37\linewidth]
        {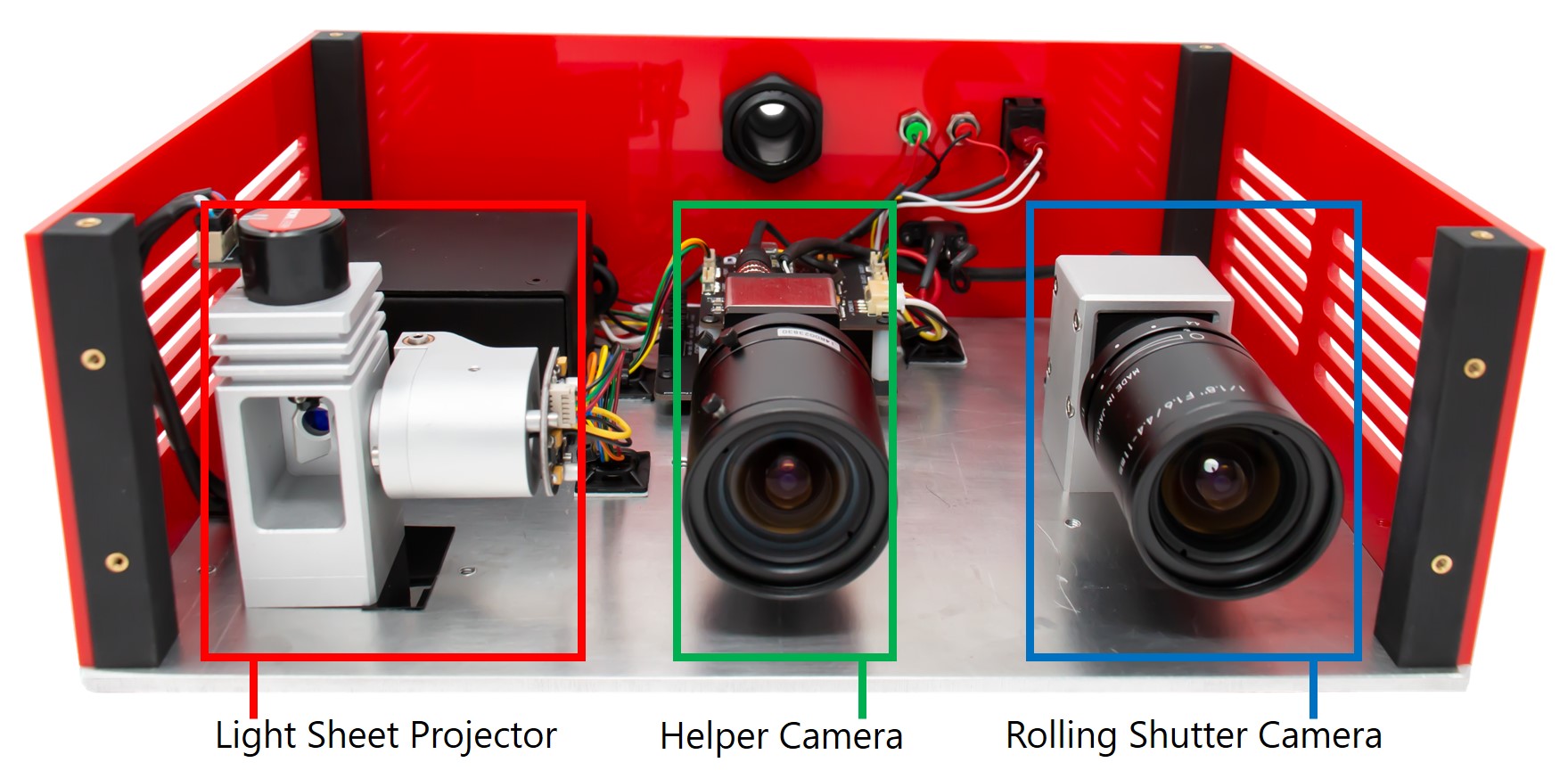}
        \label{fig:prototype}
    }
    \subfloat[\small Working principle of PLC ]{
        \includegraphics[width=0.45\linewidth]{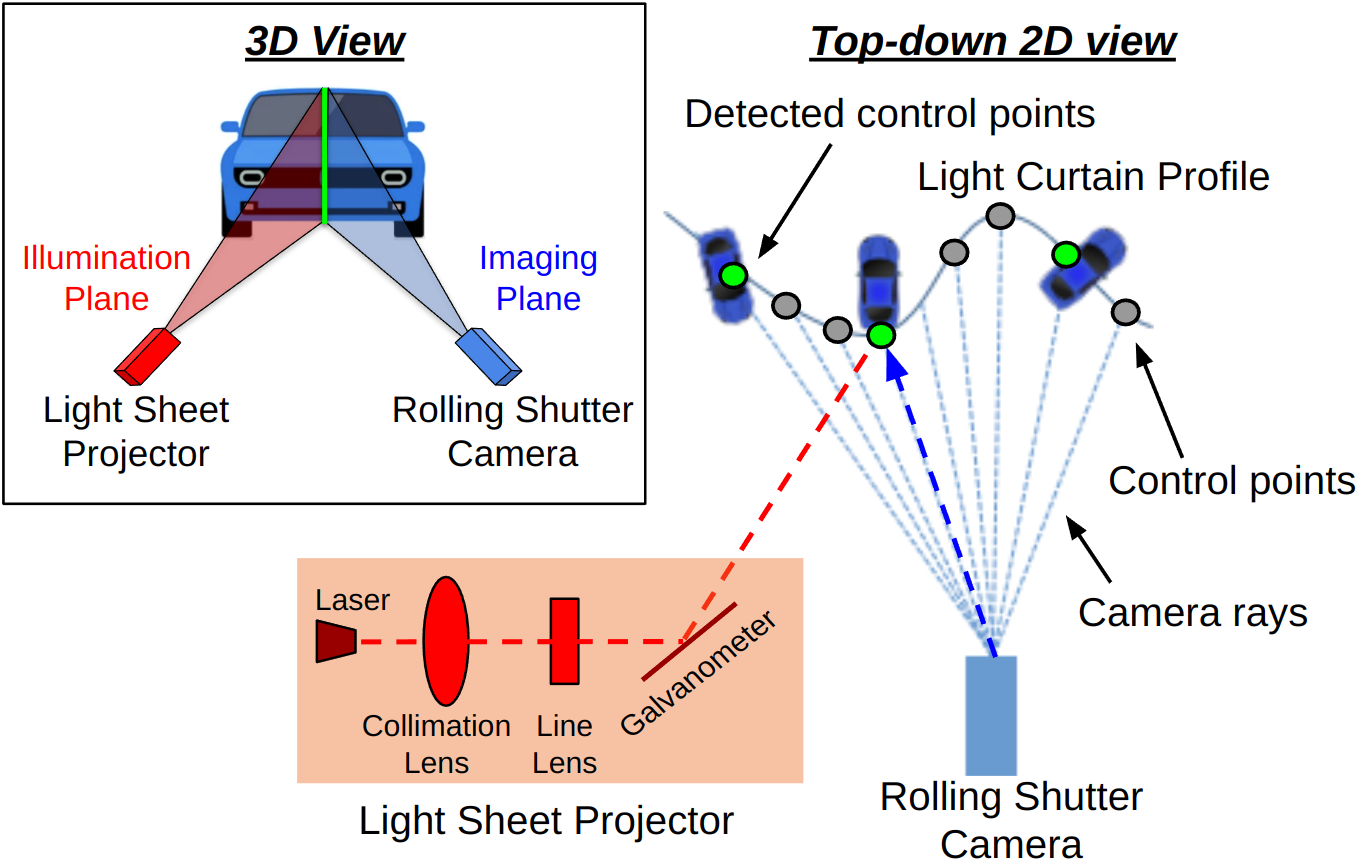}
        \label{fig:working-principle}
    }}
    \caption{(a) Our PLC prototype consists of a near-infrared (NIR) light sheet laser reflected by a rotating galvomirror, an NIR rolling shutter camera, and an additional RGB helper camera for visualization. (b) The light sheet laser rotates in synchrony with the rays of the rolling shutter camera. Only the line (green) at the intersection of the illumination plane (red) from the laser and the imaging plane (blue) of the camera is sensed at a given instant. By controlling the rotation of the laser, the green line can be made to follow a user-specified surface. Therefore, the PLC is able exclusively image 3D points on the user-specified surface at a high resolution. Figures adapted from \cite{bartels2019agile, ancha2020eccv}.}
    \vspace{-1em}
\end{figure*}

\section{Introduction}\label{sec:intro}
    \noindent Despite several advances in factory automation using robots, human operators are still required during the manufacturing process for operations that involve unpredictable behavior (e.g., fabric, rubber) or fine-tuned sensitivity (e.g., electronics assembly). To enable seamless collaboration between humans and robots in manufacturing processes, it is necessary to develop robust safety systems that protect the collaborative environment and optimize the overall production cycle \cite{matheson2019human, gualtieri2021emerging}. Such systems should incorporate advanced sensors, intelligent perception and control algorithms, and clear communication protocols.

Current safety systems can be broadly categorized into two approaches --- (1) fixed barriers and (2) vision-based collision avoidance \cite{robla2017working, buerkle2023towards}. Fixed barriers, such as fences, fixed light curtains, and time-of-flight (ToF) sensors, physically demarcate the robot's workspace and isolate it from human activity.
While fixed barriers are simple to construct or implement, and are computationally efficient, they limit collaborative work and reduce flexibility and performance in shared spaces (e.g., they prevent the robot from crossing the barrier even if no obstacle is present).
Other collision avoidance approaches utilize vision systems such as cameras \cite{ozkahraman2021design}, LiDARs \cite{zheng2021laser}, and 3D time-of-flight sensors \cite{pasinetti2018development, intel_veo} to dynamically adapt the robot's behavior based on its surroundings. This enables a higher degree of collaboration and flexibility but typically require multiple devices and powerful computational resources (multi-core CPUs and GPUs) for processing the raw sensor streams.

Programmable light curtains (PLCs) \cite{wang2018programmable, bartels2019agile} are a recently developed controllable 3D sensor that images points only along a user-specified ruled surface (called a `curtain'). These curtains, which are projected by a steerable sheet laser, can be adapted in real-time to image desired objects or regions of interest at a high resolution. The sensor has an ideal working range of up to $30$ meters, is robust even in the presence of ambient light, and is relatively inexpensive (the lab-built prototype costs about $\$1$K - $2$K).
Prior works have successfully used PLCs for various tasks in robotics such as object detection \cite{ancha2020eccv}, depth estimation \cite{raaj2021exploiting}, navigation \cite{ancha2023rss}, and safety envelope estimation \cite{ancha-se}.

In this work, we use PLCs to build a flexible safety monitoring system for industrial robots. PLCs are used to detect the presence or absence of objects around robots by creating virtual `safety curtains'
that tightly envelop the robot and adapt to the robot's configuration as it moves.
When an obstacle penetrates the safety curtain, the intrusion is detected and the robot is immediately commanded to stop.
Simultaneously, we also use light curtains to scan the scene and produce a dense 3D reconstruction of the workspace.
In addition, we show how PLCs can be optimally placed i.e. ``instrumented" in the workspace to maximize the visibility coverage of the robots. The resulting system is relatively inexpensive compared to traditional laser-based safety systems (\$$1$K-$2$K vs \$$50$K-$100$K).
Furthermore, a single PLC can monitor multiple robots within its field-of-view.
Therefore, our method easily scales to many robots using only few PLCs sensors.

In this paper, we make the following contributions towards building a PLC-based safety monitoring system:
\begin{itemize}
    \item We develop an optimization procedure to best position (instrument) multiple PLCs in a workspace to maximize visibility coverage of a given set of robots (\autoref{sec:instrumentation}).
    \item We develop algorithms for \textit{dynamic safety curtains} that tightly envelop (moving) robots at all times and detect intrusions (\autoref{sec:safety-curtain-design}, \autoref{sec:intrusion-detection}).
    \item We show how safety curtains can be interleaved with curtains that sweep across the scene to produce high-resolution 3D scene reconstructions (\autoref{sec:3d}).
    \item We evaluate our safety system in a real-world manufacturing testbed environment to monitor four robot arms using two PLCs.
    We evaluate the accuracy of our system to detect various object types, and conduct a detailed analysis of its latency and response time to safely stop a robot under intrusion (\autoref{sec:expts}).
\end{itemize}

\section{Related Work}\label{rtd}
    {
\newcommand{\emphasis}[1]{#1}
\newcommand{\hthickness}{1.1pt}
\newcommand{\vthickness}{1.1pt}
\newcommand{\NA}{\it \texttt{Not Applicable}}
\newcolumntype{I}{!{\vrule width \vthickness}}
\newcolumntype{|}{!{\vrule width 0.8pt}}
\newcolumntype{:}{!{\vrule width 0pt}}

\newcommand{\dcell}[2]{{#1}}

\setlength{\tabcolsep}{2pt}

\begin{table}[b!]
  \centerline
  {
  \begin{tabular}{Ic|c|c|c|c|cI}
   \Xhline{\hthickness}
    \textbf{Technology} & \makecell{\bf Cost} & \makecell{\bf Response\\\bf time} & \makecell{\bf Power\\\bf consumption} & \makecell{\bf Compute\\\bf Needs} & \makecell{\bf Program-\\\bf mability}\\
    \Xhline{\hthickness}
    \makecell{Multi Beam \\LiDAR \cite{altuntas2023review}} & \$\$\$ & $50$ ms & $35$ W & High & High\\
    \hline
    \makecell{3D ToF \cite{pasinetti2018development, intel_veo}} & \$\$ & $100$ ms & $100$W-$1$kW & High & High\\
    \hline
    \makecell{Cameras \\ \cite{ozkahraman2021design, yang2018multi, mohammed2017active}} & \$ & $\sim$$40$ ms & $30$-$150$W & High & High\\
    \hline
    \makecell{Ultrasonic \\Sensors \cite{digikey2022ultrasonic, tong2021ultrasonic}} & \$ & $20$-$240$ ms & $<$$5$W & Low & Low\\
    \hline
    \makecell{Fixed light \\curtains \cite{keyence}}& \$\$ &$7$-$20$ ms & $<$$5$W & Low & Low\\
    \hline
    \makecell{Prog. light \\curtains (Ours)} & \$ & $20$-$40$ ms & $10$W & Low & High\\
    
    \Xhline{\hthickness}
  \end{tabular}
  }
  \caption{
    Comparison of prior robot safety monitoring technologies for collaborative workspaces across various dimensions. Power consumption includes the computation power.
  }
  \label{table:related-work}
\end{table}
}

    \subsection{Robot Safety Monitoring}
Conventional methods for safety monitoring designate a fixed robot workspace.
Human workers are prevented from intruding this workspace either using physical fence barriers or fixed light curtains \cite{keyence}.
As robot environments become increasingly collaborative, there is a need for adaptive vision-based systems.
\autoref{table:related-work} summarizes prior technologies that can be used for robot safety monitoring in collaborative workspaces. Some advanced methods such as LiDAR~\cite{altuntas2023review, podgorelec2023lidar}, cameras~\cite{ozkahraman2021design,yang2018multi,mohammed2017active, rosenstrauch2018human} and 3D time-of-flight (ToF) sensors~\cite{pasinetti2018development, intel_veo} provide high programmability but require heavy compute and power resources. %
\\ 
In systems involving LiDAR, RGB(D) cameras, or 3D ToF sensors, there's a need for significant processing. These systems have to separate intrusions from the image/point cloud and/or use object detection models. For example, with 3D ToF, processing is necessary to identify the robot workspace, detect moving objects, and calculate the minimum distance between them to decide on actions like stopping or slowing down. Additionally, using multiple cameras to capture and merge point clouds adds to the power and computational requirements.

While relatively simpler solutions such as ultrasonic sensors~\cite{digikey2022ultrasonic, tong2021ultrasonic} and fixed light curtains have the potential to provide low latencies and low power systems, they are not programmable. Vogel et al.~\cite{project-cam} conceptualized a programmable safety system using a visible-light projector and camera, but without a real implementation using robots. The safety system presented and implemented in this paper provides a highly programmable and accurate solution, requires minimal computation, and provides a low response time while keeping the overall cost low.

\subsection{Programmable Light Curtains} 
PLC was first introduced in \cite{wang2018programmable, bartels2019agile} as a controllable depth sensor that can sense along any vertically-ruled surface at high-resolution. Chan et. al. \cite{chan2022holocurtains} recently upgraded the sensor to be able to sense along any arbitrary-shaped 2D surface using a holographic projector. We use the original sensor, based on a line laser and rolling-shutter camera, in our setup as it can adapt its sensing surface in real-time.

Prior works have used PLCs for perception tasks such as 3D object detection\cite{ancha2020eccv} and monocular depth sensing~\cite{raaj2021exploiting}. Ancha et. al.~\cite{Ancha-RSS-21} used PLCs for computing dynamic safety envelopes in unknown environments, which is closest to our work. Ancha et. al. \cite{ancha2023rss} extended light curtains for velocity estimation of dynamic obstacles in unknown environments.
The aforementioned tasks fall in the category of \textit{active perception} where the next light curtain to be placed depends on previous sensor measurements.
Our work differs from active perception by focusing on the specific setting of indoor safety monitoring for manufacturing robots where the positions and motion of robots are \textit{known} apriori. This removes the need for active sensing and forecasting for curtain placement, while also increasing the need for real-time tracking with high accuracy.

\section{Background on Prog. Light Curtains}\label{bkg}
    Programmable light curtains (PLCs) \cite{wang2018programmable, bartels2019agile} are a recently-developed controllable depth sensor that images only along a specified vertically-ruled 2D surface in the environment. The
sensor contains two main components: a rolling-shutter near-infrared (NIR) camera
and a rotating sheet NIR laser (see \autoref{fig:prototype}). The rolling-shutter camera activates one pixel column at a time, and we refer to the top-down projection
of the imaging plane corresponding to each pixel column as a “camera ray” (see \autoref{fig:working-principle}). A 2D control point is selected on each camera ray (shown as gray and green circles). A controllable galvo-mirror rotates the laser light sheet in synchrony with the rolling shutter camera to point the laser sheet at the control point corresponding to the currently active pixel column. 3D scene points that lie at the surface
of this intersection between the projected light sheet and the image plane (called `curtain') get imaged by the camera (shown as green circles). The set of control points completely determine the shape of the light curtain and subsequently the objects that are imaged. These control points form the input to the PLC. The control points can be specified at each timestep to place a curtain in each imaging iteration of the camera. Additionally, an RGB camera is used as a `helper camera' to visualize the projected light curtains as shown in \autoref{fig:teaser}. Please refer to Bartels et. al. \cite{bartels2019agile} for more details about PLCs such as its hardware specification and performance under various lighting conditions.

\begin{figure}[t!]
    \definecolor{light-red}{RGB}{240, 128, 128}
    \centering
    \includegraphics[scale=0.5]{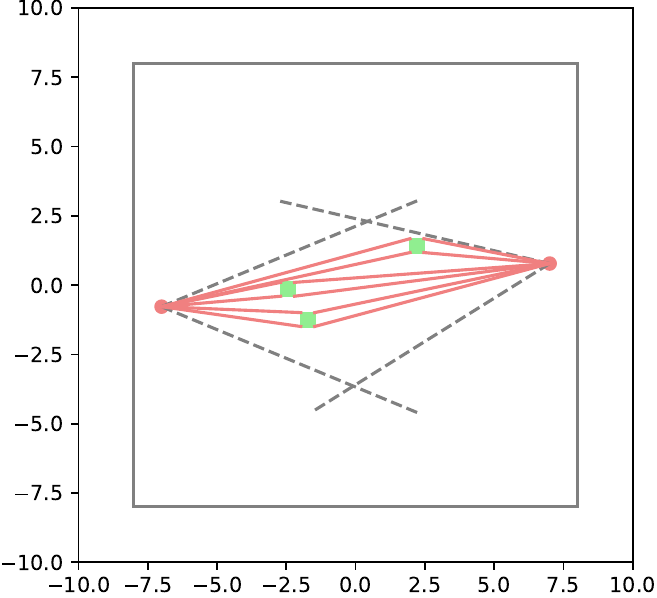}
    \caption{Optimized configuration of PLCs (red circles) for an $8m \times 8m$ layout containing three robots (green squares). Our algorithm computes the configuration that maximizes the sum of the angles subtended by the robots at the PLCs (rays denoted by {\color{light-red}\textbf{---}}, field-of-view denoted by {\color{gray}\textbf{{-}{-}{-}}}) over many samples of joint PLC configurations.}
    \label{fig:instrument-eg}
    \vspace{-1em}
\end{figure}    

\section{Method}\label{sec:method}
    In this work, we study the problem of building an inexpensive safety monitoring system using PLCs. Towards this goal, we first address the question of how to place (i.e. \textit{instrument}) the sensors in the robot workspace or factory floor to maximize the safety coverage of the robots. Then, we discuss how to design the shape of the projected light curtains so that they can act as a dynamic `safety shield' around the robot and detect object intrusions. Finally, we discuss how the PLCs can be used to obtain a full 3D reconstruction of the workspace.

\subsection{PLC instrumentation to maximize robot coverage}
\label{sec:instrumentation}

For a given layout and positions of robot arms in a work-space, we begin by determining the optimal positions and orientations of the PLCs that can safely envelope all the robots.
Instead of manually specifying the positions and orientations guided by intuition or trial-and-error, we take a principled algorithmic approach as follows.

We represent each robot and PLC in $2$D from the top-down view, as shown in \autoref{fig:instrument-eg}. Each robot is represented by a $4$-sided polygon (shown in green), and the location of the PLC device is represented by a point (shown as red circles). The four vertices of each robot polygon are provided as input to our algorithm and remain fixed, while the $2$D pose ($x, y, \theta$) of each PLC needs to be estimated. The continuous search space for  2D pose is reduced to a discretized grid of poses. 

Na\"ive brute-force search over this grid yields an exponential time-complexity of $\mathcal{O}(N^{3M})$, where $N$ is the number of discretized $x$, $y$ and $\theta$ values searched over, and $M$ is the number of PLCs. The computationally complexity is prohibitively large. Instead, our approach is to generate $n$ \textit{uniformly random samples} from this search space, inspired by RANSAC \cite{fischler1981random}.

After we sample a $2$D pose for each PLC, we determine the robots contained within each PLC's field-of-view. For every robot within the field-of-view, we compute the \textit{angle subtended} by the edge formed by the two closest vertices of the robot polygon at the PLC. Importantly, we only consider a vertex if it has not already been observed by another PLC in order to avoid double-counting the coverage of the robot.

The \textit{sum of angles} subtended by the observed edges at every PLC is used as a measure of coverage quality for every sampled configuration. Higher scores are given to configurations where all vertices of a robot are covered. We repeat this process for each of the $n$ sampled PLC configurations and return the configuration with the highest sum of subtended angles. The number of PLCs required to completely envelope all robot edges can be determined by simply running the algorithm for an increasing number of PLCs, starting from 2, until all the visible edges of the robots are covered.
The full algorithm is described in \autoref{alg:plc_pose_estimation}. \begin{algorithm}
  \caption{PLC Instrumentation Algorithm}
  \label{alg:plc_pose_estimation}

  \KwIn{Robot polygons, 2D pose grid, number of sample iterations $n$.}
  \KwOut{Optimal 2D poses of $M$ PLCs, $C$.}
  Initialize $C=\emptyset$\;
  \For{$i = 1$ to $n$}{
    Initialize set of observed polygon edges $S=\emptyset$\;
    Randomly sample $M$ 2D poses $(x, y, \theta)$ from grid\;
    Determine robots within PLC's field-of-view\;
    \ForEach{PLC}{
    \ForEach{robot $\in$ field-of-view(PLC)}{
      Determine the two closest corners of robot from PLC\;
      \If{any corner has not been observed by another PLC}{
      Compute angle subtended by their edge at PLC\; 
      Add edge and angle to $S$\;
      }
      }
      \If{all corners of robot have been observed}{
      Add 10 to sum of angles in $S$\;
      }
    }
    \If{sum of angles in $S$ $>$ sum of angles in $C$}{
      Update $C$ with current PLC poses and obtained sum\;
    }}
  \Return{$C$}\;
\end{algorithm}
\vspace{-1em}

The result of this algorithm for an example containing three robot arms and two PLC sensors is shown in \autoref{fig:instrument-eg}. The algorithm was run for $10^4$ iterations on a $50\times50\times20$ grid. Further results are discussed in \autoref{sec:results}.

\subsection{Safety curtain design}
\label{sec:safety-curtain-design}

We design safety curtains from each PLC based on the known poses of the robot arms in the scene. For this, we first obtain the joint positions from the robot arms' joint encoders, compute their 3D locations using forward kinematics, and then project them into the 2D top-down view in the frame of reference of the closest PLC. Since the number of joint positions are usually small ($6$-$8$), we also use the positions of some pre-defined virtual joints from the robot's geometry. We then compute a 2D convex hull in the 2D top-down view enclosing all of the 2D robot points using the Quickhull algorithm \cite{qhull}. Finally, to compute the control points of the curtain, we perform 2D ray-tracing for each camera ray of the PLC in the 2D plane containing the convex hull. The points of intersections of each ray with the hull determine the control points and hence the shape of the safety curtain. A small additional offset is added to the control points to ensure the curtains don't intersect with the robots while being sufficiently close to the robot's surface. 
This computation is repeated at each time step as each robot moves around its workspace, ensuring that the safety curtains dynamically adapt and closely track the robots' motion in real-time.

\subsection{Intrusion Detection}
\label{sec:intrusion-detection}

If an object intersects the safety curtain, it is detected as a high-intensity return on the raw image of the PLC. Once an object detection is observed, we determine which robot(s) it corresponds to based on the 3D location of the detection. This robot is then instructed to stop immediately. Interference from other sensors may also occasionally cause intensity returns. Since these interference returns can usually last between 2 to 4 frames, we stop a robot only if there is a continuous intensity return for more than 4 consecutive frames. In situations where utmost safety is prioritized over efficiency, the robots can be temporarily stopped whenever an intensity return is received, regardless of whether it originates from an object or interference. Additionally, detections due to interference exhibit a distinct pattern. One direction of future work could be to train a small on-device neural network to determine if the intensity return is caused by interference or an object.

\subsection{Scene Reconstruction}\label{sec:3d}
In addition to placing safety curtains, we also obtain high-resolution depth maps by sweeping the scene with fixed-shape (or random) curtains. We merge the points detected by each curtain in the sweep to output a full depth reconstruction. Each intensity return has an associated depth map as each pixel corresponds to a known camera-ray -- laser sheet intersection. These intensity returns from a full sweep can then be merged together into a single intensity image by storing the maximum intensity value at each pixel location across the sweep of images. This merged intensity image is then backprojected to a full 3D point cloud using the known depth value and the camera's intrinsics. Since the PLC can project multiple light curtains at the camera's frame rate of $45$-$60$ Hz, random curtains (meant for reconstruction) can be interleaved with the safety curtains (meant for collision avoidance). The point clouds generated from each sensor are sent to a workstation where they are processed and registered together using the iterative closest point (ICP) algorithm \cite{icp}.
The resulting 3D map of the static environment could be further used to detect other dynamic objects such as mobile robots or people navigating in the workspace.

\section{Experiments}\label{sec:expts}
    \subsection{Setup}

\begin{figure}
    \centering
    \begin{subfigure}{0.45\textwidth}
        \centering
        \includegraphics[width=1\textwidth]{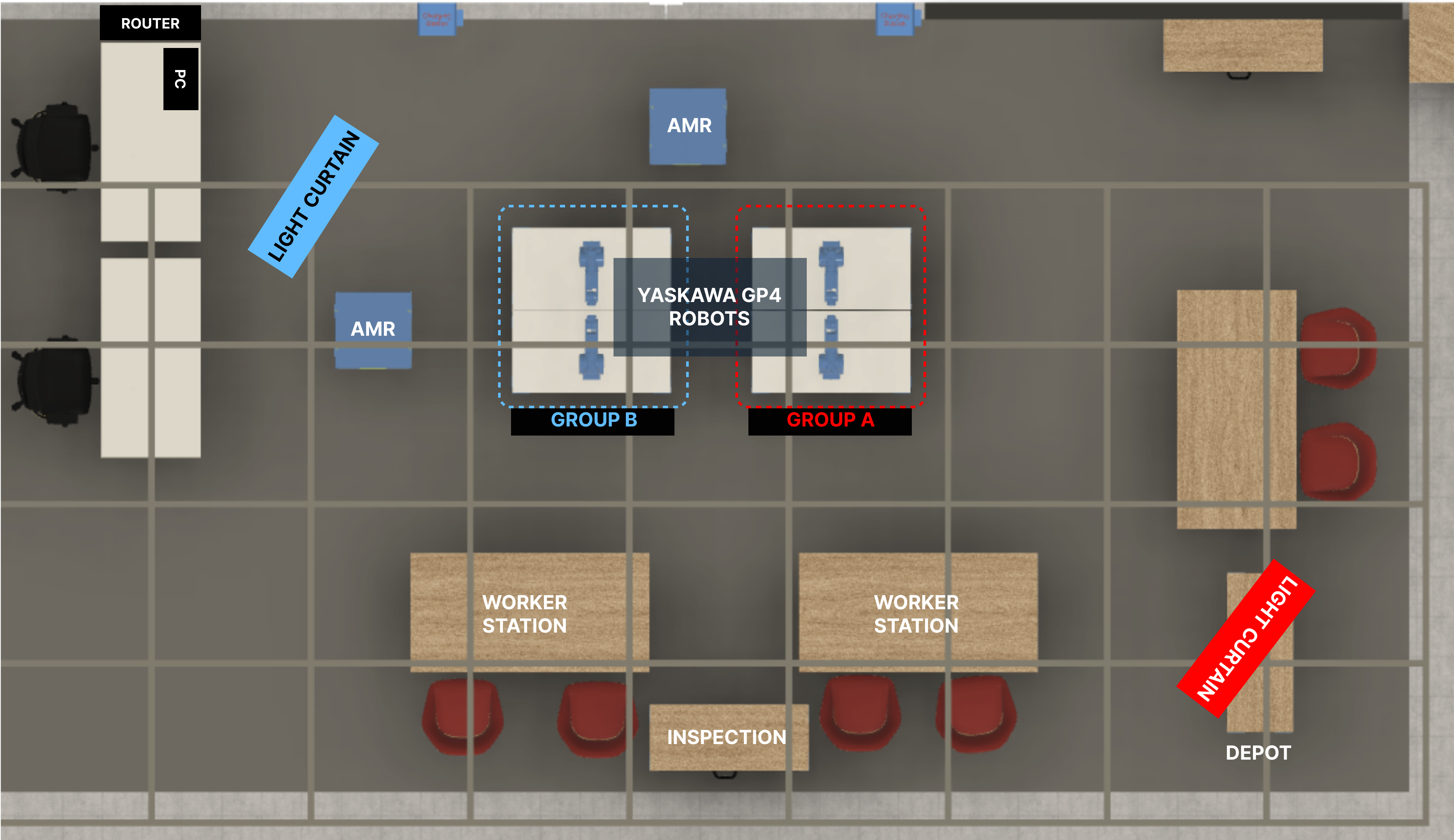}
        \caption{\textit{Testbed layout.} Four robot arms are arranged together in pairs on workstations that are $0.7$ m from the ground. Two PLCs (in blue and red) are mounted on an $80/20$ grid suspended from the ceiling at a height of $3.35$ m.}
        \vspace{1em}
    \end{subfigure}
    \begin{subfigure}{0.45\textwidth}
        \centering
        \includegraphics[width=\linewidth]{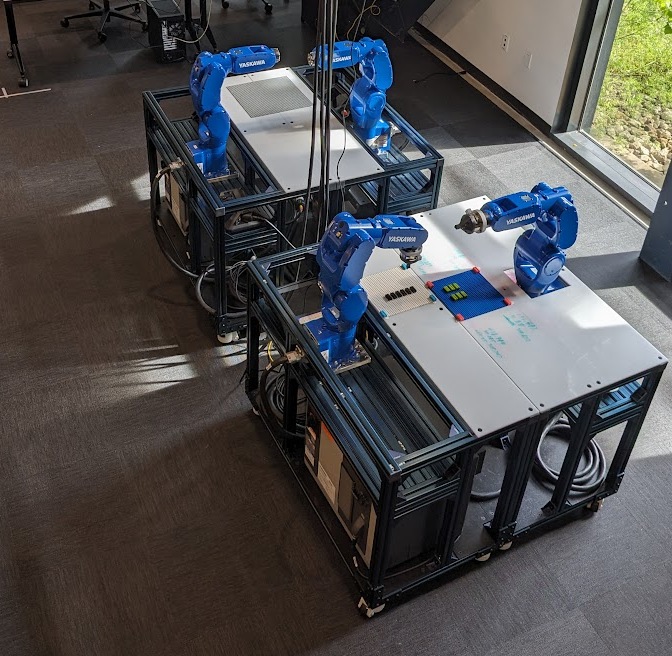}
        \caption{Four Yasakawa\textsuperscript{\textregistered} robot arms observed from the point of view of the red PLC. 
        }
    \end{subfigure}
    \caption{Testbed area used in our experiments.}
    \label{fig:testbed}
    \vspace{-1em}
\end{figure}

\begin{figure}
    \centering
    \includegraphics[width=\linewidth]{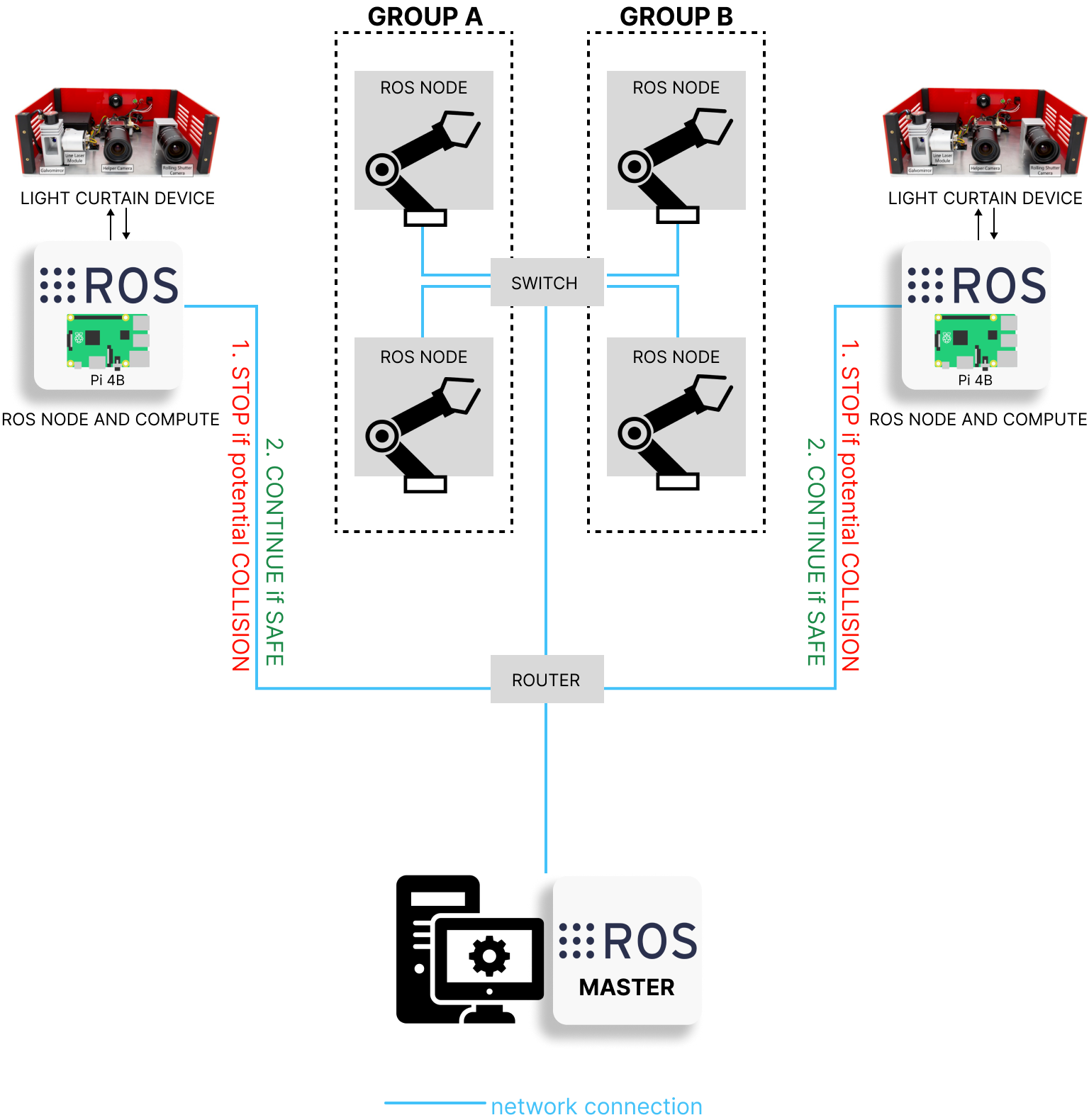}
    \caption{Network setup used in the testbed. The PLC sensors are connected to the robots over ROS using Ethernet cables. The robots are monitored and controlled for safety directly by the PLC sensors. The workstation is used for visualization and additional control.}
    \label{fig:system-diagram}
    \vspace{-2em}
\end{figure}

\noindent\textit{\textbf{Testbed}}: We demonstrate the capabilities of our safety monitoring system in a manufacturing testbed shown in \autoref{fig:testbed}. The testbed consists of two pairs of Yasakawa\textsuperscript{\textregistered} GP4 robotic arms arranged in a square layout, with each pair collaborating to assemble and disassemble a Lego\textsuperscript{\textregistered} model. Each robot is equipped with its own control PC which is also connected to a central workstation over ROS. There are two autonomous mobile robots (AMRs) that move material between the arm robots and the worker stations. We mount two downward-facing PLCs on a $80$/$20$ grid attached to the ceiling at a height of $3.35$ m from the ground. Each PLC has a Raspberry-Pi\textsuperscript{\textregistered}~4 onboard which executes the code for designing and imaging the light curtains. Including the RPi, each PLC draws up to 10W power. The dimensions of the testbed are $9.3\times5.9$ m\textsuperscript{2} and all inter-process communication is carried over Ethernet to minimize latency. The pose between each robot and PLC is calibrated with an eye-on-base calibration procedure using Apriltags \cite{hand-eye}. Due to the high engineering effort and/or cost of implementing other safety systems in the testbed, we limit our experiments and analysis to only our system.\\

\begin{figure}
    \centering
    \includegraphics[width=\linewidth]{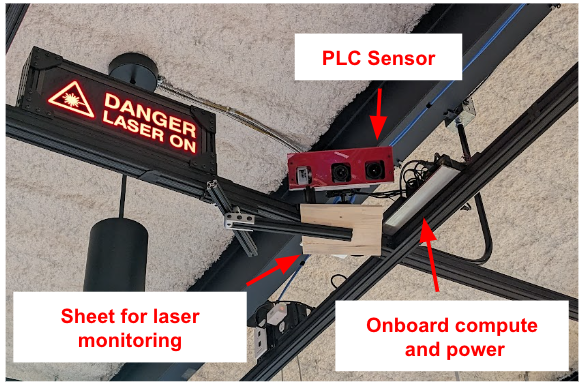}
    \caption{Red PLC sensor mounted in the testbed with on-board compute and power, and a laser safety mechanism.}
    \label{fig:plc-annotate}
    \vspace{-0.1in}
\end{figure}

\noindent\textit{\textbf{Eye Safety}}: Eye safety requirements limit the maximum power at which the laser on the PLC can be operated \cite{achar, ansi}, which limits the maximum range of the sensor. In our testbed, we operate the laser at 30\% of its maximum power (1W), which corresponds to a maximum eye-safe distance of 16.80 cm. This distance is well beyond the reach of people (with a height of at-most 9 feet) during regular operation. As further precaution, we also monitor the output of the sensor to ensure that the galvanometer mirror is rotating and that the laser's power level is maintained. We do so by attaching a white planar sheet as an extension to the sensor at a short distance within its field-of-view (shown in \autoref{fig:plc-annotate}) without introducing any additional circuitry into the prototype sensor. This sheet is scanned by the PLC every second and if the intensity return from it does not match the expected return, the mirror is deemed malfunctioning and the sensor is immediately turned off. The returned intensity values are also monitored to ensure that the laser's output power is constant and is not fluctuating.

\subsection{Results}\label{sec:results}

\begin{figure*}[t!]
    \centerline{
    \subfloat[\small]{
        \includegraphics[width=0.24\linewidth]
        {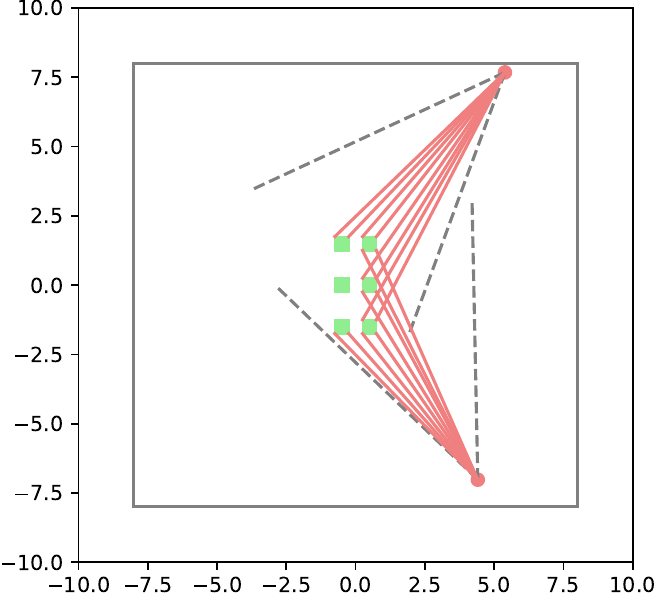}
        \label{fig:cube}
    }
    \subfloat[\small]{
        \includegraphics[width=0.24\linewidth]{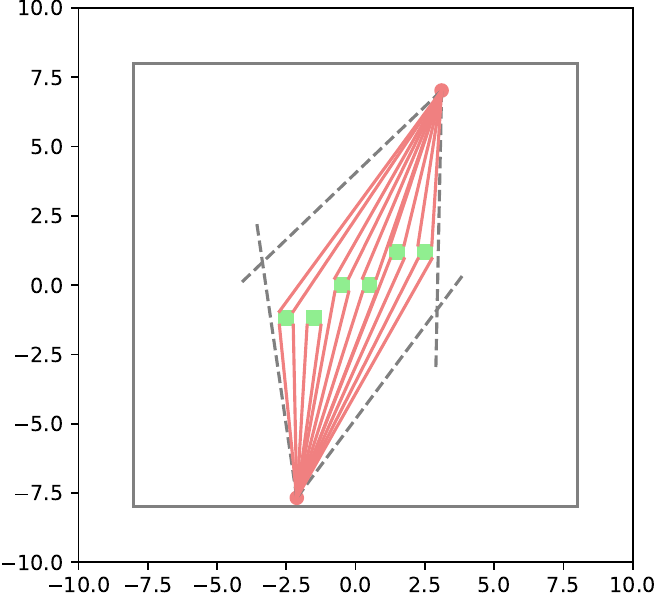}
        \label{fig:stagg}
    }
    \subfloat[\small]{
        \includegraphics[width=0.24\linewidth]{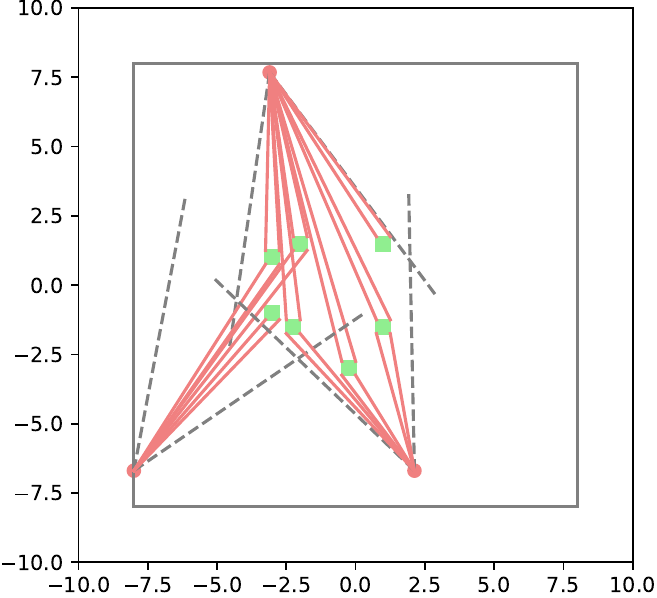}
        \label{fig:rand}
    }
    \subfloat[\small]{
         \includegraphics[width=0.24\linewidth]{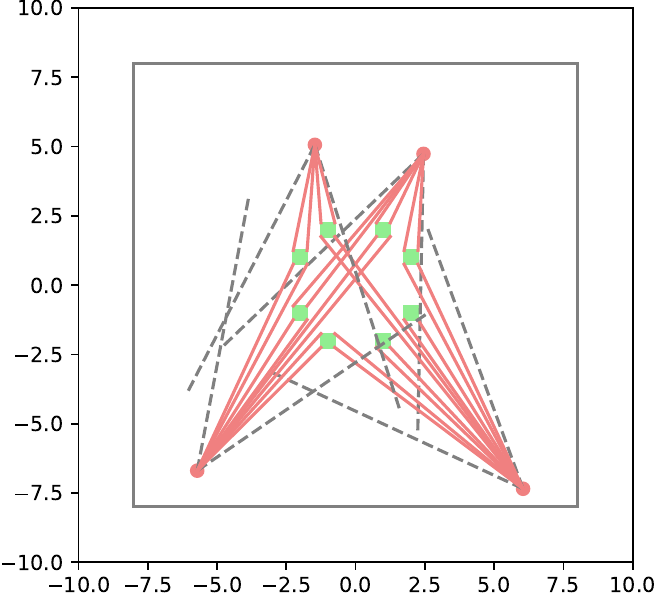}
        \label{fig:octo}
    }}
    \caption{Results of our random sampling based instrumentation algorithm on four robot configurations (after $10^6$ iterations each). We define coverage to be the percentage of total vertices of the robot polygons (green squares) that are visible. Six robots arranged in a grid can be $67$\% covered using two PLCs. \textbf{(b)} When the same six robots are staggered, they can be covered $100$\% covered using only two PLCs. \textbf{(c)} Seven robots in shown random positions can be $86$\% covered using three PLCs. \textbf{(d)} Eight robots arranged as an octagon can be $87.5$\% covered with four PLCs.}
    \label{fig:instrument-results}
\end{figure*}

\begin{figure}
    \centerline{
    \subfloat[\small]{
        \includegraphics[width=0.55\columnwidth]{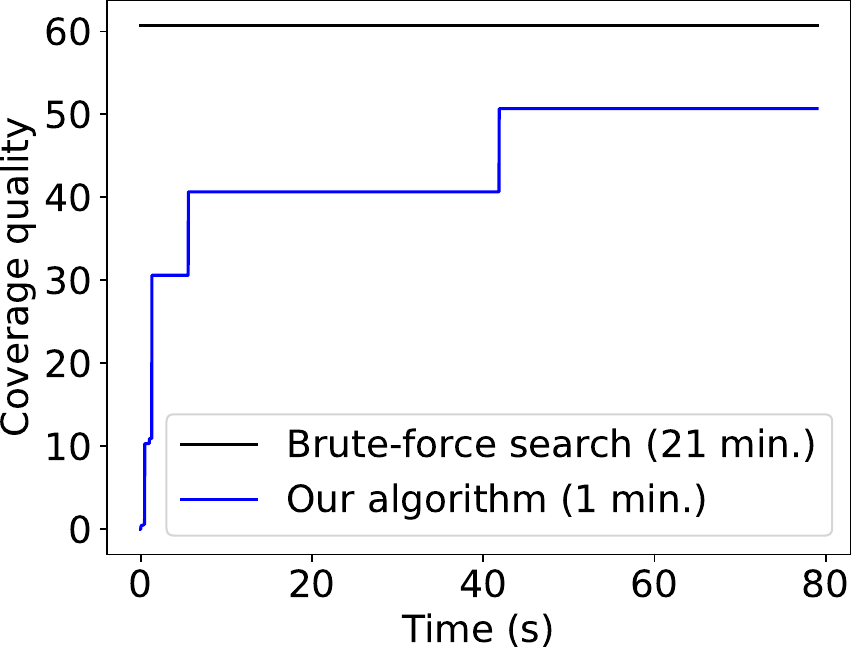}
        \label{fig:working-principle}
    }}
    \vspace{0.5em}
    \centerline{
    \subfloat[\small]{
        \includegraphics[width=0.48\columnwidth]{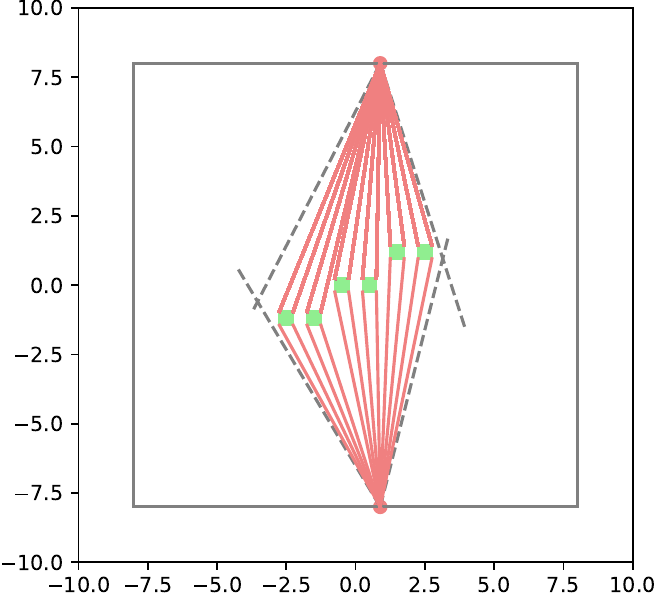}
        \label{fig:working-principle}
    }
    \subfloat[\small]{
        \includegraphics[width=0.48\columnwidth]{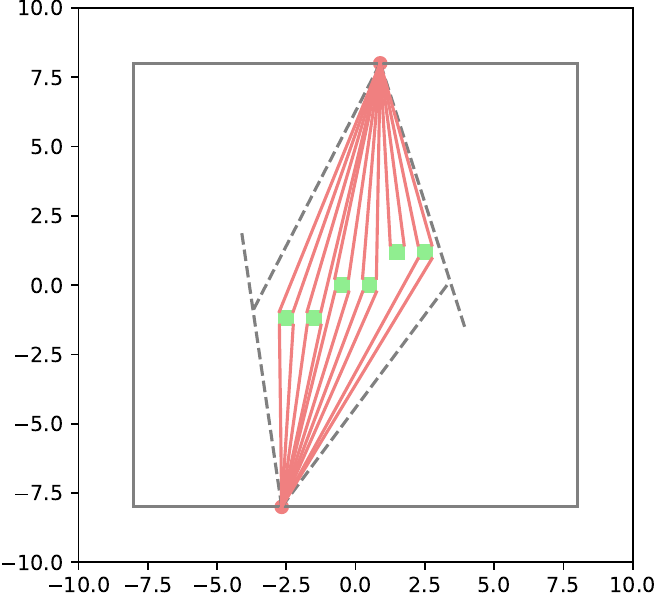}
        \label{fig:working-principle}
    }}
    \caption{A comparison of our random sampling based instrumentation algorithm against brute-force search with two PLCs on a $10\times10\times20$ grid size. Note that for larger grid sizes and more number of PLCs, brute-force search becomes infeasible. \textbf{(a)} Our algorithm achieves a comparable coverage quality in a fraction of the time ($1$ minute $19$ seconds vs. $21$ minutes $09$ seconds). \textbf{(b)} Configuration found using brute-force search. \textbf{(c)} Configuration found using ours.}
    \label{fig:instrument-results2}
\end{figure}

\noindent\textit{\textbf{Instrumentation}}: The two PLCs in our testbed are positioned according to results from our algorithm described in \autoref{sec:instrumentation}. Unlike regular laser-based safety sensors that typically can monitor only one robot each, our two PLCs can monitor all four robots as well as the full workspace area for people and other mobile robots.

We provide additional results of our algorithm on four robot layouts (grid, staggered, randomized, octagon) in \autoref{fig:instrument-results}. The algorithm was run on a grid size of $50\times50\times20$ for $10^6$ iterations and took approximately $15$ minutes for each layout on a single Intel Xeon W2123 CPU core. It was run for $5\times10^6$ iterations on the octagon layout with $4$ PLCs which took about $3$ hours to complete. We consider a robot to be \textit{fully enveloped} if all its four corners are observable. Clearly, the layout of the robots determines the number of PLCs required to safely envelope all the robots. \autoref{fig:stagg} shows that the more staggered the robots are, the fewer is the number of PLCs required. We also compare our algorithm against a brute-force search approach, on a smaller search grid of $10\times10\times20$, in \autoref{fig:instrument-results2}. Our algorithm approaches the performance of brute-force search in a significantly smaller amount of time and scales easily to more number of PLCs and larger grid sizes.\\

\noindent\textit{\textbf{Safety curtains}}: The blue and red PLCs receive the joint positions of the group A and group B robots respectively over ROS at $40$ Hz. These positions are used to compute, in real-time, the safety curtains required to envelope the robots in their current pose. The safety curtains are re-computed for every frame and projected to follow the robots' movement, ensuring that the robots are always enveloped. This tracking behaviour of the sensors is visualized in the video on our \href{\website}{website}. Unlike regular laser-based safety systems which typically block the area around the robots, our safety curtains follow the robots more tightly which enables fence-less human-robot collaboration.\\

\noindent\textit{\textbf{Intrusion detection}}: When a safety curtain intersects an object, the reflected laser light is captured directly by the NIR camera. These intrusions are detected at the optical level by design and are hence very accurate. Visualizations of such a detection are shown in \autoref{fig:teaser}. Objects of varying sizes from a full-sized human to small parts of the hand can be quickly detected. For objects that are of dark colors, the laser operating power would need to be increased to detect them as dark objects tend to reflect a smaller fraction of incident light. When an intrusion is detected, the corresponding robot is immediately commanded to stop. The robot resumes its motion once the intruding object disappears. The accuracy of the setup was evaluated by throwing different sized objects into the robot's workspace. We report the frequencies of detections in \autoref{table:accuracy}.\\

\begin{minipage}{\linewidth}
    \begin{minipage}[b]{\linewidth}
        \small
        \centering
        \begin{tabular}{cc}
            \toprule
            \textbf{Objects} & \textbf{Detection frequency} \\
            \midrule
            Human Walking & 20 / 20 ~(100\%) \\
            Cap &  20 / 20 ~(100\%) \\
            Paper Ball & 18 / 20 ~(\phantom{0}90\%) \\
            LEGO 4x2 Block & 15 / 20 ~(\phantom{0}75\%) \\ 
            \bottomrule
        \end{tabular}
    \captionof{table}{Number of accurate detections over 20 trials of different objects thrown into the robot's workspace.}
    \label{table:accuracy}
    \vspace{2em}
    \end{minipage}
    \newline
    \begin{minipage}[b]{\linewidth}
        \centering
        \includegraphics[scale=0.2]{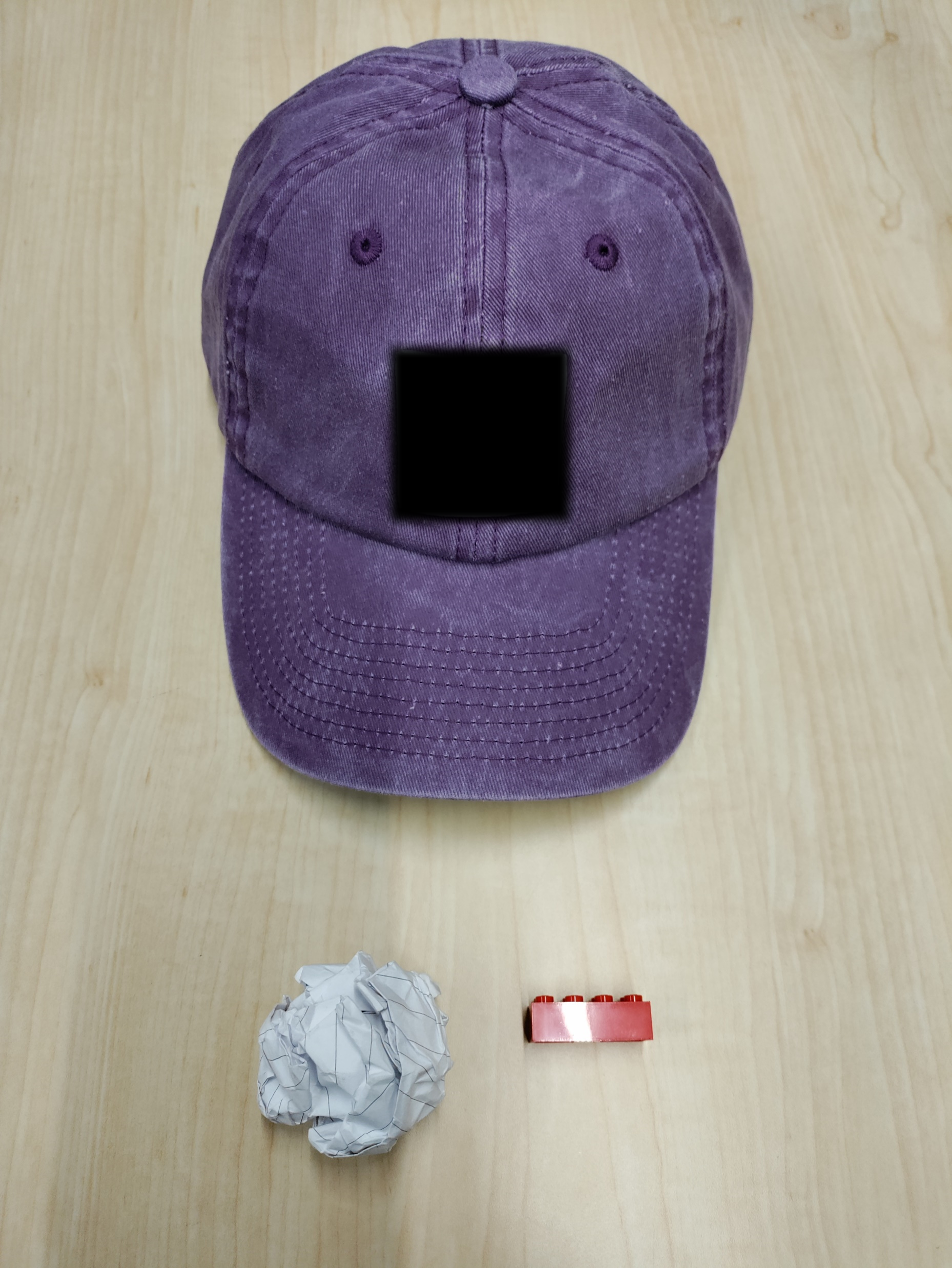}
        \label{fig:objects}
        \captionof{figure}{Objects of different sizes thrown into the robot's workspace to evaluate detection accuracy.}
       \vspace{1em}
    \end{minipage}
\end{minipage}

\noindent\textit{\textbf{Latency}}: 
Apart from accuracy, the overall system response time is also important to ensure safety. The PLC sensor is fast and it can in principle operate at rates up to the imaging rate of the rolling-shutter camera ($30$-$45$Hz). In practice, it is limited by the time taken to compute the curtains, which varies according to the processor used. On the RPI-4 used in our setup, the imaging rate of the planar and random curtains is $24$ Hz (or response time of 42 ms). The imaging rate of the dynamic safety curtains that are designed and imaged according to the robot's motion is $7$ Hz. The computationally expensive step is the ray-hull intersections, which implemented as unoptimized Python code takes $130$ms on average on Raspberry Pi 4B. We quantify the latencies involved in the rest of the system by repeatedly projecting fixed planar curtains and recording the time taken for the robot to stop upon intrusion, over many trials. Results are reported in \autoref{table:latencies}. Like the imaging rate, the time taken to issue the STOP command can also be improved with a better processor (for e.g., $9$ ms on an i9 processor). The time taken for the robot to stop completely (approx. $300$ ms) is a limitation of the robot arm driver.\\

{
\newcommand{\emphasis}[1]{#1}
\newcommand{\hthickness}{1.1pt}
\newcommand{\vthickness}{1.1pt}
\newcommand{\NA}{\it \texttt{Not Applicable}}
\newcolumntype{I}{!{\vrule width \vthickness}}
\newcolumntype{|}{!{\vrule width 0.8pt}}
\newcolumntype{:}{!{\vrule width 0pt}}

\newcommand{\dcell}[2]{{#1}}

\setlength{\tabcolsep}{2pt}

\begin{table}[th!]
  \centerline
  {
  \begin{tabular}{IcIc|cI}
   \Xhline{\hthickness}
    \textbf{Event} & \makecell{\textbf{Time event occurs}\\ {(fixed planar curtains)}} & \makecell{\textbf{Time event occurs}\\ {(dynamic safety curtains)}}\\
    \Xhline{\hthickness}
    Object intruded &  0 ms & 0 ms \\
    \hline
    Intrusion detected &  42 ms & 143 ms\\
    \hline
    STOP issued &  91 ms & 192 ms\\
    \hline
    Robot stopped &  374 ms & 475 ms\\
    \Xhline{\hthickness}
  \end{tabular}
  }
  \caption{
    Median times for events in the intrusion detection pipeline. Fixed planar curtains image at 24 Hz, whereas dynamic safety curtains image at 7 Hz due to the convex hull computation required at every frame.
    \vspace{-1.2em}
  }
  \label{table:latencies}
\end{table}
}

\begin{figure}[h]
    \centering
    \begin{subfigure}{0.45\textwidth}
        \centering
        \includegraphics[width=0.8\linewidth]{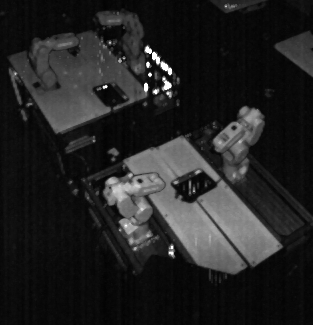}
        \caption{Merged intensity returns from the blue PLC.}
        \label{fig:mintensity}
        \vspace{1em}
    \end{subfigure}
    \begin{subfigure}{0.45\textwidth}
        \centering
        \includegraphics[width=0.8\linewidth]{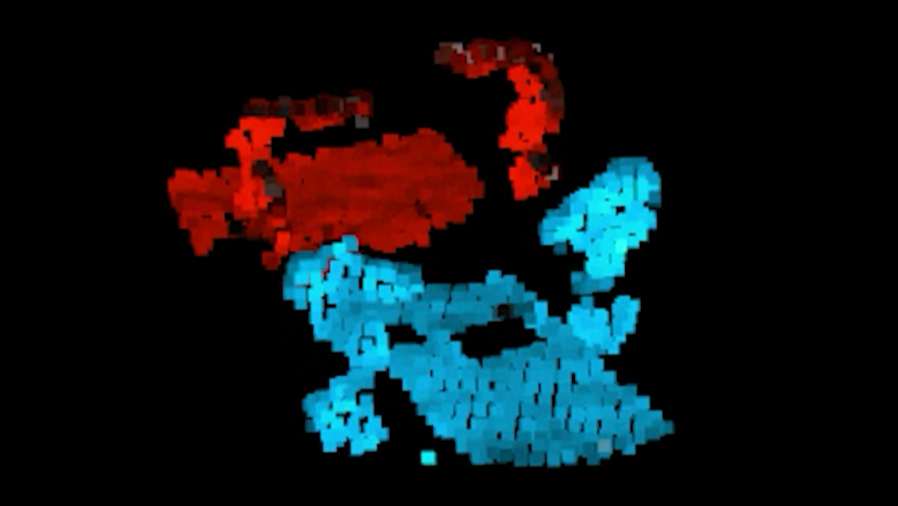}
        \caption{Merged point cloud from the blue and red PLCs.}
        \label{fig:pointcloud}
        \vspace{1em}
    \end{subfigure}
    \caption{\textit{Pointcloud reconstruction.} The testbed area is scanned using planar curtains by the blue and red PLCs, and their intensity returns are processed and registered together to obtain a reconstruction in 3D.}
    \vspace{-1em}
\end{figure}

\noindent\textit{\textbf{3D Reconstruction}}: Apart from projecting safety curtains, the PLC sensors also sweep fixed-shape (planar) curtains at regular intervals ($1$ cm) to obtain a full 3D reconstruction of the testbed area.
The returned intensity images are merged into a single intensity image (shown in \autoref{fig:mintensity}) and its corresponding pointcloud reconstruction is obtained as described in \autoref{sec:3d}. The pointclouds from both PLCs are then filtered and registered together using ICP on the central workstation with the calibrated poses of the PLCs used as initialization.
The resulting pointcloud for the four robots is visualized in \autoref{fig:pointcloud}.

\section{Conclusion}\label{sec:discuss}
    We presented a new inexpensive safety monitoring system using programmable light curtains for collaborative manufacturing. The system is flexible and can be used for monitoring arbitrary configurations of dynamic robots as well as the full 3D scene, while easily scaling to many robots.
Future work would be to incorporate the robot motion plan and trajectory forecasting of obstacles into the pipeline to avoid recomputing safety curtains at each frame.
Such forecasting would fully leverage PLC's active sensing capabilities and enable new active safety monitoring systems that are able to pre-empt dangerous situations instead of responding passively. Likelihood-based sampling methods can be used to further speed-up the instrumentation search algorithm. Current limitations of the system include inaccurate depth sensing at large distances (greater than 8 m) due to curtain thickness \cite{bartels2019agile} which can be alleviated using probabilistic depth models \cite{raaj2021exploiting}.
Another drawback is less sensitivity to dark objects, which could be alleviated by adapting the laser intensity online to the sensed objects.

\textbf{Acknowledgements}: This research is supported by the CMU Manufacturing Futures Institute, made possible by the
Richard King Mellon Foundation. KR would like to thank Swaminathan Gurmurthy and Anurag Ghosh for helpful discussions.

\hypersetup{
    urlcolor=black
}

\footnotesize{
\bibliographystyle{IEEEtran}
\bibliography{references}
}

\addtolength{\textheight}{-12cm}   %

\end{document}